\documentclass[sigconf]{acmart}

\usepackage{booktabs}
\usepackage{multirow}
\usepackage{amsmath}
\usepackage{graphicx}
\usepackage{pgfplots}
\pgfplotsset{compat=1.18}
\usepackage{placeins}

\setcopyright{acmcopyright}
\copyrightyear{2026}
\acmYear{2026}
\acmConference[RecSys '26]{16th ACM Conference on Recommender Systems}{September 2026}{Minneapolis, Minnesota}
\acmBooktitle{Proceedings of the 16th ACM Conference on Recommender Systems (RecSys '26),
              September 2026, Minneapolis, Minnesota}
\acmDOI{10.1145/XXXXXXX.XXXXXXX}
\acmISBN{978-1-4503-XXXX-X/26/09}

\title{Attribute-Conditioned Multimodal Slot Factorization\\
       for Controllable Fashion Retrieval}

\author{Najmeh Forouzandehmehr}
\affiliation{%
  \institution{Walmart Global Tech}
  \city{Sunnyvale}
  \state{California}
  \country{USA}
}
\email{najmeh.forouzandehmehr@walmart.com}

\author{Topojoy Biswas}
\affiliation{%
  \institution{Walmart Global Tech}
  \city{Sunnyvale}
  \state{California}
  \country{USA}
}
\email{topojoy.biswas@walmart.com}

\author{Evren Korpeoglu}
\affiliation{%
  \institution{Walmart Global Tech}
  \city{Sunnyvale}
  \state{California}
  \country{USA}
}
\email{evren.korpeoglu@walmart.com}

\author{Kannan Achan}
\affiliation{%
  \institution{Walmart Global Tech}
  \city{Sunnyvale}
  \state{California}
  \country{USA}
}
\email{kannan.achan@walmart.com}


\begin{abstract}
Fashion retrieval often requires satisfying multiple attributes at once,
such as category, color, pattern, and demographic. Monolithic embeddings
mix these signals into a single vector, making attribute-specific control
difficult at retrieval time. Many existing semantic-ID methods provide
discrete item codes, but these codes are typically optimized as item-level
or residual addresses and do not expose named, independently controllable
attribute slots.

We introduce \textbf{MM-slotgate}, a multimodal slot encoder that factorizes
Fashion-CLIP text and image embeddings into four named attribute slots.
Each slot learns its own text--image gate, so visually grounded attributes
such as color and pattern can rely more on image evidence, while
taxonomy-oriented attributes such as category and demographic can remain
more text-driven.

On H\&M, using a combined slot-similarity and slot-logit retrieval score,
MM-slotgate achieves 0.7566 macro ConstraintSatisfied@10, outperforming
equal-weight multimodal fusion (0.7142) and fCLIP text-only retrieval
(0.4755). The largest gain is on color, which improves from 0.321 to
0.889 (\(+0.568\) absolute), as the learned color gate assigns 57.4\%
weight to image evidence. The learned gates are interpretable without
modality supervision: color is image-leaning, category is text-leaning,
and pattern and demographic lie near the middle.

The resulting slots also remain controllable: linear probes show no
measured excess leakage beyond the label-correlation baseline, and
quantized slot codes support targeted intervention, including a
\(15.3\times\) lift for color. These results suggest that controllable fashion
retrieval benefits from typed, attribute-conditioned multimodal slots rather
than either a single global embedding or opaque item-level semantic IDs.
\end{abstract}


\begin{document}

\maketitle


\section{Introduction}
\label{sec:intro}

Fashion retrieval systems must simultaneously satisfy constraints across
multiple orthogonal attributes: a query for ``women's patterned summer
dresses'' requires matching category, demographic, and pattern axes at once.
Standard retrieval pipelines encode queries as monolithic dense vectors,
making it difficult to selectively emphasize one attribute at query time
without constructing an entirely new query.

Supervised slot factorization improves constraint-satisfaction retrieval
by decomposing item representations into named attribute slots, each
optimized for a different semantic dimension. However, slot factorization
alone does not determine which input modality each slot should use. This is
especially important in fashion: color and pattern are often primarily
visual, while category and demographic are often more explicit in product
text and catalog taxonomy. A text-only slot encoder therefore underuses
visual evidence, while a single fixed text--image blend applies the same
modality mixture to every attribute. We instead argue that modality fusion
should be attribute-conditioned: each slot should learn how much to rely on
text versus image evidence.

We address this gap with \textbf{MM-slotgate}: a multimodal slot encoder
that fuses Fashion-CLIP (fCLIP)~\cite{fashionclip2023} text and image
embeddings through per-slot learnable gates.
Each slot $s$ receives a scalar gate $g_s = \sigma(a_s)$, trained
end-to-end, that controls the text-versus-image blend entering that
slot's VQ bottleneck.
Rather than prescribing a fixed modality ratio, the model learns that
color is best resolved by images ($g_\text{color}=0.426$, image-leaning),
while category and demographic are resolved by text
($g_\text{categ}=0.545$, $g_\text{demo}=0.509$, text-leaning)—without
any modality supervision.

We make three contributions:
\begin{enumerate}
  \item \textbf{MM-slotgate architecture}: a per-slot learnable gate over
        fCLIP text and image embeddings that learns attribute appropriate
        modality allocation end-to-end.
        A graceful missing-image fallback ensures full catalog coverage.
  \item \textbf{Strong empirical results}: MM-slotgate achieves 0.7566
        macro ConstraintSatisfied@10 on H\&M, surpassing equal-weight
        MM-global fusion (0.7142, $+$5.9\% relative) and fCLIP text-only
        (0.4755, $+$59.1\%).
        The color constraint improves from 0.321 (fCLIP-text) to 0.889
        ($+$0.568 absolute), the largest single-attribute gain observed.
  \item \textbf{Gate analysis and ablation}: Learned gates converge to
        attribute-appropriate modality preferences without modality
        supervision (color image-leaning, category text-leaning) and
        outperform fixed equal-weight fusion on macro CS@10
        (0.7566 vs.\ 0.7516, $+$0.005 absolute).
        A shuffled-image negative control confirms that actual visual
        content, not merely co-training with an extra input branch, drives the gain.
\end{enumerate}


\section{Related Work}
\label{sec:related}

\subsection{Semantic IDs and Vector Quantization}

Semantic ID methods~\citep{tdm_recsys_2018,joint_tdm_2019,rajput2023}
represent items as structured discrete codes for retrieval and generation.
TIGER~\citep{rajput2023} learns item IDs via residual quantization and
generates the next item's code sequence autoregressively; subsequent work
uses these IDs as ranking features~\citep{semantic_ids_ranking_2024} or
enforces hierarchical structure~\citep{seater_2024}.
VQ-Rec~\citep{vqrec_2023}, LETTER~\citep{letter_2024}, and
TokenRec~\citep{tokenrec_2025} learn discrete tokens for sequential and
LLM-based recommendation.
Recent work also studies disentangled or tag-aligned semantic
IDs~\citep{hidvae_2025}.

Our approach differs in the semantics assigned to each code.
Existing methods learn item-level or residual addresses whose meaning
emerges implicitly; we instead define four named attribute slots—pattern,
color, category, demographic each with its own codebook.
This makes the representation directly addressable: one slot can be
weighted or intervened on independently, and the semantic role of each
code is fixed by supervised alignment rather than discovered implicitly.

\subsection{Multimodal Fashion Retrieval}

CLIP~\citep{radford2021} and its fashion-domain adaptation
fCLIP~\citep{fashionclip2023} provide strong monolithic embeddings by
aligning image and text in a shared space.
These embeddings improve per-attribute retrieval precision but blend all
attributes into a single vector, making selective emphasis difficult.
MM-slotgate uses fCLIP as a backbone but replaces the global embedding
with four attribute-specific slots, each with its own learned modality gate.

\subsection{Controllable and Disentangled Retrieval}

Compositional retrieval~\citep{vo2019} edits the \emph{query} representation
at inference time using relative attribute feedback.
Our approach is complementary: we decompose \emph{index-time item}
representations, enabling attribute intervention without recomputing any
embedding.
Unsupervised disentanglement ($\beta$-VAE~\citep{higgins2017}) does not
guarantee alignment with task-relevant attributes; we use supervised slot
alignment and evaluate disentanglement relative to a label correlation
baseline~\citep{han2017}.


\section{MM-Slotgate: Multimodal Slot Encoder}
\label{sec:model}

\begin{figure*}[t]
  \centering
  \includegraphics[width=\textwidth,interpolate=true]{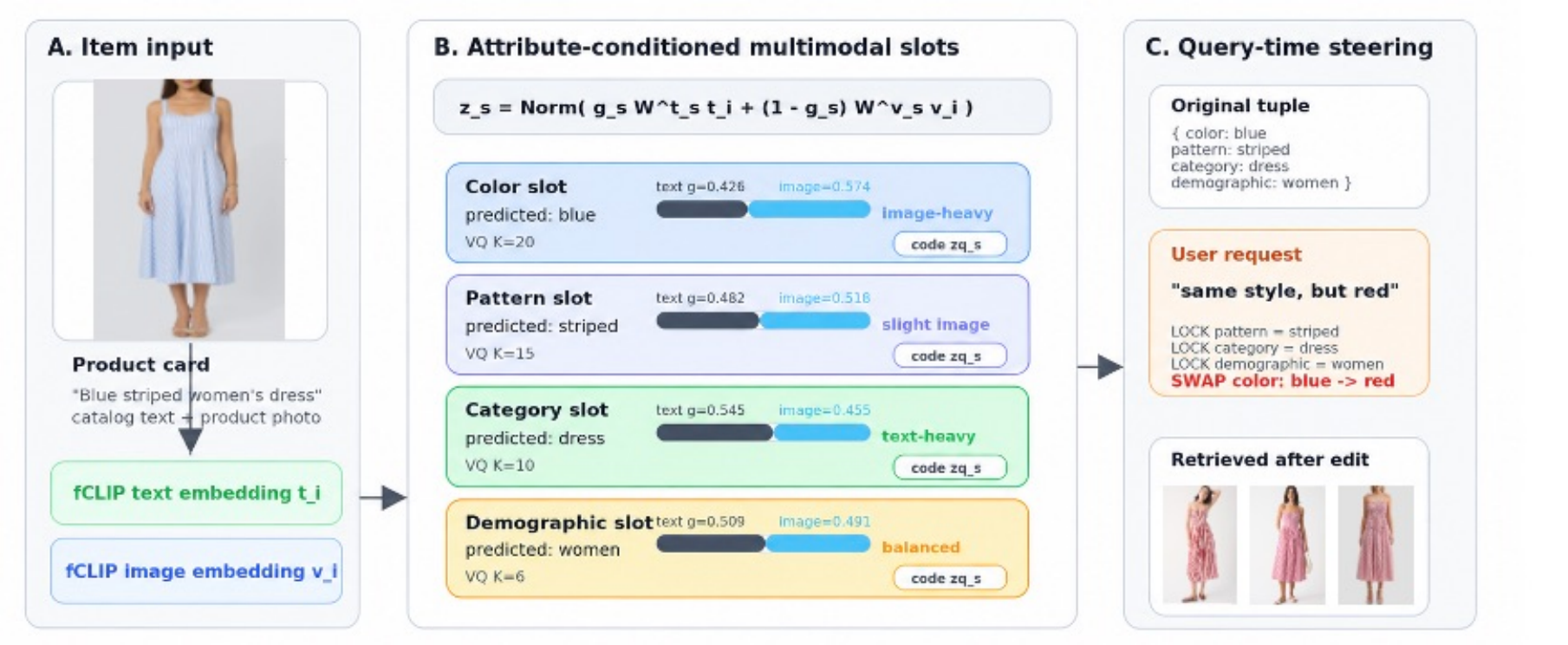}
  \caption{MM-slotgate decomposes a fashion item into four named multimodal slots.
  Each slot learns its own text--image gate before vector quantization.
  Continuous slots support weighted retrieval; quantized slot codes support
  targeted intervention, illustrated by changing only the color slot from
  blue to red.}
  \label{fig:mm-slotgate-steerability}
\end{figure*}

Figure~\ref{fig:mm-slotgate-steerability} gives the retrieval and intervention
intuition before we define the model formally.

MM-slotgate has three stages.
First, each item is embedded with Fashion-CLIP text and image encoders.
Second, each semantic slot learns its own text--image mixture and produces a
continuous slot vector $\mathbf{z}_{i,s}$.
Third, each slot is quantized into a slot-specific codebook, yielding
$\mathbf{z}^q_{i,s}$.
Retrieval uses the continuous slots $\mathbf{z}$, while intervention uses the
quantized slots $\mathbf{z}^q$.

\noindent\textbf{Inputs:}

We encode each item $i$ with the Fashion-CLIP ViT-B/32
backbone~\cite{fashionclip2023}, producing a 512-dim text embedding
$\mathbf{t}_i$ (product name + description) and a 512-dim image embedding
$\mathbf{v}_i$ (product photograph).
A binary indicator $m_i \in \{0,1\}$ flags image availability
(${\approx}95\%$ on H\&M).
\emph{Note}: ``demographic'' denotes product target audience
(men's, women's, kids'), not a user attribute.

\smallskip
\noindent\textbf{Per-slot projection and gating:}

For each slot $s \in \{\text{pattern, color, category, demographic}\}$:
\begin{equation}
  \mathbf{h}^t_{i,s} = \mathrm{LN}(W^t_s\,\mathbf{t}_i),\quad
  \mathbf{h}^v_{i,s} = \mathrm{LN}(W^v_s\,\mathbf{v}_i),
  \label{eq:proj}
\end{equation}
where $W^t_s, W^v_s \in \mathbb{R}^{128 \times 512}$ are slot-specific
projections with Layer Normalization.
A per-slot scalar $a_s$ (initialized to $0$, learned end-to-end) defines the
text weight $g_s = \sigma(a_s) \in (0,1)$; thus $g_s$ is the text weight and
$1{-}g_s$ is the image weight for slot $s$.
Values near $0.5$ give equal fusion; $g_s{>}0.5$ is text-leaning;
$g_s{<}0.5$ is image-leaning.

\medskip
\noindent\textbf{Fusion with missing-image fallback:}

For items with an available product image, each slot mixes its text
and image projections according to the learned gate; for items without
an image, the slot uses the text projection only:

\begin{equation}
  \tilde{\mathbf{z}}_{i,s}
    = m_i\!\left(g_s\,\mathbf{h}^t_{i,s}
               + (1{-}g_s)\,\mathbf{h}^v_{i,s}\right)
    + (1{-}m_i)\,\mathbf{h}^t_{i,s},
  \label{eq:fusion}
\end{equation}
followed by L2-normalization: $\mathbf{z}_{i,s} = \mathrm{Norm}(\tilde{\mathbf{z}}_{i,s}) \in \mathbb{R}^{128}$.
When $m_i=0$, the slot falls back to $\mathrm{Norm}(\mathbf{h}^t_{i,s})$,
preserving full catalog coverage.

\smallskip
\noindent\textbf{Vector quantization:}

Each slot has an EMA codebook of $K_s$ unit-norm codewords
($K_s \in \{15, 20, 10, 6\}$ for pattern, color, category, demographic).
A straight-through estimator~\cite{bengio2013} passes gradients through the
discrete assignment:
$\mathbf{z}^q_{i,s} = \mathbf{z}_{i,s} + \mathrm{sg}(\mathbf{e}_{i,s} - \mathbf{z}_{i,s})$.
Dead entries (EMA count $< 0.05$) are restarted from random training items.

\smallskip
\noindent\textbf{Training loss:}

\begin{equation}
  \mathcal{L} = \sum_{s=1}^{4}\Bigl[
    \underbrace{\beta\|\mathbf{z}_{i,s} - \mathrm{sg}(\mathbf{e}_{i,s})\|^2}_{\text{commitment}}
    + \underbrace{\lambda_a\,\mathcal{L}_{\mathrm{CE}}(\text{logits}_s,\,y_s)}_{\text{alignment}}
  \Bigr]
  + \lambda_\perp \mathcal{L}_{\mathrm{orth}},
  \label{eq:loss}
\end{equation}
with $\beta{=}0.25$, $\lambda_a{=}5.0$, and $\lambda_\perp{=}2.0$.
$\mathcal{L}_\mathrm{orth}$ penalizes off-diagonal elements of the
slot Gram matrix, discouraging slot collapse.
Optimization details are given in Section~\ref{sec:retrieval}.

\smallskip
\noindent\textbf{Combined retrieval:}

At query time, the score for item $i$ given constrained query $q$ is:
\begin{equation}
  \text{score}(q,i) = \cos\!\bigl(\mathbf{w}(q),\,\mathbf{w}(i)\bigr)
    + \alpha \!\sum_{s \in \mathcal{C}} \log \hat{P}(y_s = c^*_s \mid i),
  \label{eq:retrieval}
\end{equation}
where $\mathbf{w}(\cdot)$ is slot-weighted concatenation (constrained slots
$3\times$, others $1\times$, L2-normalized; \emph{Wsel-$z$}),
$\hat{P}$ from alignment logit heads, and $\alpha^*$ tuned by 5-fold CV.
Equation~\eqref{eq:retrieval} is called the \emph{combined} score.


\section{Experimental Setup and Retrieval Evaluation}
\label{sec:retrieval}

We first ask whether slot-specific multimodal fusion improves constraint satisfaction.

\paragraph{Dataset.}
We evaluate on the H\&M Personalized Fashion Recommendations
catalog~\cite{hm2022kaggle}, using a 50K-item subset with product text,
product images, and four attribute labels: pattern, color, category, and
demographic.
We use a 90/10 split: 45K items for training and 5K held out for evaluation.
Image availability is approximately 95\%; items without images use the
text-only fallback defined in Eq.~\eqref{eq:fusion}.

\paragraph{Training protocol.}
All MM-slotgate variants are trained for 80 epochs on the 45K-item training
split with AdamW (learning rate $10^{-3}$, cosine decay), EMA codebook decay
$0.99$, and dead-code restart.

\paragraph{Retrieval protocol.}
We evaluate CS@$n$ on the 5K held-out split at $n{=}10$.
For each constraint set, we sample 500 random query items and retrieve from
the held-out gallery excluding the query itself.

\paragraph{Metric.}
\textit{ConstraintSatisfied@$n$} (CS@$n$): fraction of top-$n$ retrieved items
where \emph{all} constrained slots match the query's ground-truth labels.
\textit{Chance} is $\prod_{s \in \mathcal{C}} \sum_c p(y_s=c)^2$.

\paragraph{Methods.}
\begin{itemize}
  \item \textbf{CLIP-text}: combined retrieval score on CLIP ViT-B/32
        text embeddings with slot encoder.
  \item \textbf{fCLIP-text-only}: combined retrieval on fCLIP text embeddings
        (no image input; text-only MM-slotgate variant with $g_s \approx 1$).
  \item \textbf{MM-global}: fuses text and image \emph{before} the slot encoder,
        using the precomputed $\mathrm{Norm}(0.5\,\mathbf{t}_i + 0.5\,\mathbf{v}_i)$
        as input—a single global mixture with no per-slot gating.
        (Distinct from \textbf{fixed\_half} in Table~\ref{tab:ablation},
        which keeps separate per-slot text/image projections but freezes
        each gate at $g_s{=}0.5$.)
  \item \textbf{MM-slotgate (ours)}: combined retrieval with per-slot
        learned gates (Eq.~\eqref{eq:retrieval}).
  \item \textbf{MetaFilter+fCLIP$^\dagger$} (oracle):  the item
        catalog is filtered to items whose attribute labels exactly match
        the query's constrained slots, then ranked by fCLIP cosine
        similarity. Requires knowing the answer labels at retrieval time
        and is therefore unreachable in practice; represents the best
        result a perfect attribute filter could achieve.
\end{itemize}

\paragraph{Results.}
Table~\ref{tab:main} shows macro CS@10.
MM-slotgate (0.7566) outperforms MM-global (0.7142) by $+$5.9\% relative.
Gate ablation analysis (\S\ref{sec:gates}) shows that learned per-slot
allocation slightly outperforms fixed equal-weight fusion while providing
interpretable attribute-specific modality allocation (0.7566 vs.\ 0.7516).
Both multimodal methods far exceed fCLIP text-only (0.4755), with gains of
$+$58.8\% (MM-slotgate) and $+$50.2\% (MM-global) relative to the text
baseline—demonstrating that image information is essential for
constraint-satisfying retrieval on H\&M.

Table~\ref{tab:constraints} reports per-constraint CS@10 for all
three methods across the nine constraint sets. MM-slotgate leads on
every row, with the largest gains on color-containing constraints.
The strongest single-attribute gain is on color: fCLIP-text combined
retrieval yields 0.321, while MM-slotgate reaches 0.889
(\(+0.568\) absolute). This aligns with the learned color gate
(\(g_{\text{color}}=0.426\), image weight \(57.4\%\)): product images
provide a more direct signal for color class than product descriptions,
where lexical variants such as ``navy,'' ``midnight,'' and ``ink'' can
map to the same label class.

\begin{table}[t]
  \centering
  \caption{Macro ConstraintSatisfied@10 (H\&M, combined retrieval).
           MetaFilter$^\dagger$ is an oracle (GT labels used to filter).}
  \label{tab:main}
  \small
  \begin{tabular}{lr}
    \toprule
    Method & Macro CS@10 \\
    \midrule
    MetaFilter+fCLIP$^\dagger$  & 0.990 \\
    \midrule
    MM-slotgate (ours)           & \textbf{0.7566} \\
    MM-global fusion             & 0.7142 \\
    fCLIP-text-only              & 0.4765 \\
    CLIP-text          & 0.4640 \\
    \bottomrule
    \multicolumn{2}{l}{\footnotesize $^\dagger$ oracle upper bound; combined = Wsel-$z$ + slot logits}
  \end{tabular}
\end{table}

\begin{table}[t]
  \centering
  \caption{Per-constraint CS@10 (H\&M, combined retrieval).
           Bold = best per row.
           MM-slotgate leads on all nine constraint sets.}
  \label{tab:constraints}
  \small
  \setlength{\tabcolsep}{4pt}
  \begin{tabular}{lrrr}
    \toprule
    Constraint & fCLIP-txt & MM-global & MM-slotgate \\
    \midrule
    color              & 0.321 & 0.852 & \textbf{0.889} \\
    category           & 0.861 & 0.861 & \textbf{0.889} \\
    demographic        & 0.749 & 0.804 & \textbf{0.849} \\
    pattern            & 0.647 & 0.812 & \textbf{0.824} \\
    \midrule
    cat+demo           & 0.642 & 0.705 & \textbf{0.757} \\
    color+categ        & 0.225 & 0.670 & \textbf{0.725} \\
    color+demo         & 0.210 & 0.643 & \textbf{0.718} \\
    patt+categ         & 0.492 & 0.659 & \textbf{0.685} \\
    cat+demo+color     & 0.141 & 0.422 & \textbf{0.474} \\
    \midrule
    \textit{Macro}     & 0.4765 & 0.7142 & \textbf{0.7566} \\
    \bottomrule
  \end{tabular}
\end{table}


\section{Gate Analysis and Ablation}
\label{sec:gates}

We next examine whether the learned gates match intuitive modality needs for each attribute.

\paragraph{Learned gate values.}
Table~\ref{tab:gates} shows the per-slot gate values $g_s = \sigma(a_s)$
after 80 training epochs.
Recall that $g_s$ weights the text projection: $g_s > 0.5$ is text-leaning,
$g_s < 0.5$ is image-leaning (image weight $= 1-g_s$).

\begin{table}[t]
  \centering
  \caption{Learned gate values $g_s$ at convergence.
           Text weight $= g_s$; image weight $= 1-g_s$.
           All gates initialized to $g_s = 0.5$ ($a_s=0$).}
  \label{tab:gates}
  \small
  \begin{tabular}{lrrl}
    \toprule
    Slot       & $g_s$ & Image wt. & Modality preference \\
    \midrule
    color       & 0.426 & 0.574 & image-leaning \\
    pattern     & 0.482 & 0.518 & balanced \\
    demographic & 0.509 & 0.491 & text-leaning \\
    category    & 0.545 & 0.455 & text-leaning \\
    \bottomrule
  \end{tabular}
\end{table}

The color gate diverges furthest from $0.5$ toward the image modality,
consistent with its dominant per-constraint gain ($0.321 \to 0.889$).
Category and demographic lean text, consistent with these slots being
determined by taxonomic labeling conventions rather than visual appearance.
Pattern falls between the two: pattern names (``striped'', ``floral'') have
reasonable text coverage but benefit from visual confirmation.
Critically, these preferences emerge without any modality supervision—the
gate is trained solely to minimize alignment CE loss and commitment loss
on multi-attribute labels.

\paragraph{Gate ablations.}
Table~\ref{tab:ablation} compares four gate-mode variants on validation loss
and per-slot classification accuracy (80 epochs).

\begin{table}[t]
  \centering
  \caption{Gate ablation: macro CS@10 and color slot accuracy.
           Learned gates give the best CS@10, while fixed-half is close;
           the main benefit of learned gates is interpretable slot-specific
           modality allocation.}
  \label{tab:ablation}
  \small
  \setlength{\tabcolsep}{4pt}
  \begin{tabular}{lrrr}
    \toprule
    Mode & Macro CS@10 & Color acc & Val loss \\
    \midrule
    learned     & \textbf{0.7566} & 0.762 & 28.91 \\
    fixed\_half & 0.7516          & 0.763 & 28.77 \\
    image\_only & 0.6618          & \textbf{0.786} & 29.82 \\
    text\_only  & 0.4765          & 0.304 & 36.33 \\
    \bottomrule
  \end{tabular}
\end{table}

Learned gates produce the best macro CS@10, but the margin over fixed\_half
is small (0.7566 vs.\ 0.7516).
We therefore interpret the gate primarily as an interpretable and controllable
mechanism for attribute-specific modality allocation, rather than as the sole
source of the retrieval gain.
The large drops for text\_only (0.4765) and image\_only (0.6618) show that
both modalities are necessary: text\_only collapses on color (acc 0.304, near
chance), while image\_only degrades on category and demographic where text
provides the primary signal.

\paragraph{Negative control.}
We test whether the image branch helps because it contains the correct
product image, rather than simply because the model has an extra input.
To do this, we train the same model after randomly shuffling image
embeddings across items, so each product is paired with another product's
image. Under this shuffle, the model learns to rely less on images: all
gates become text-leaning (\(g \approx 0.57\)--\(0.63\)), and validation
loss becomes close to the text-only model (36.95 vs.\ 36.33). This suggests
that MM-slotgate benefits from correctly aligned visual content, not merely
from adding an extra image branch.

\paragraph{Representation diagnostics.}
We also run a simple diagnostic to check where color information is stored
before slot factorization. We train logistic classifiers on frozen fCLIP text,
image, and global-fusion embeddings, independently of MM-slotgate. These
classifiers are not retrieval baselines; they only measure how much attribute
information is present in each input representation. The image and global
multimodal embeddings achieve high color accuracy (0.802 and 0.789),
showing that color is strongly available in the visual branch. This matches
the learned MM-slotgate behavior: the color slot assigns more weight to the
image stream.


\section{Measuring Slot Disentanglement}
\label{sec:disentangle}

High retrieval accuracy alone is not sufficient for controllability; the slots must also avoid excess cross-slot leakage.

Multimodal fusion can, in principle, increase cross-slot information leakage
if visual features are correlated across attribute axes.
We measure this with the linear probe AUC matrix protocol from prior work.

\paragraph{Linear probe AUC matrix (MM-slotgate).}
We fit logistic regression probes from each slot's $\mathbf{z}_s$ to each
slot's ground-truth label on a held-out 10\% split.
The $4{\times}4$ AUC matrix (rows = probe target, cols = feature slot)
distinguishes own-slot encoding (diagonal) from cross-slot leakage (off-diagonal).

Table~\ref{tab:auc_mm} shows the MM-slotgate AUC matrix.
Diagonal mean AUC is $0.897$, substantially higher than the prior text-only
slot encoder ($0.798$).
The largest gains are on color ($0.641 \to 0.912$) and pattern
($0.790 \to 0.866$)—the two visually grounded slots.

Although the raw off-diagonal AUC is 0.671, this does not by itself imply
model-induced leakage because H\&M labels are correlated.
The label-prior probe, which uses only dataset co-occurrence, has off-diagonal
AUC 0.690.
MM-slotgate is slightly \emph{below} this baseline, giving excess leakage of
$-0.019$.
Thus, under linear probes, the model does not add measurable cross-slot leakage
beyond what is already present in the label structure.

\begin{table}[t]
  \centering
  \caption{Linear probe AUC matrix for MM-slotgate.
           Rows $=$ probe target, cols $=$ feature slot.
           Diagonal mean $= 0.897$; off-diagonal mean $= 0.671$;
           excess leakage $= -0.019$, indicating no measured excess leakage beyond the label-correlation baseline.}
  \label{tab:auc_mm}
  \small
  \begin{tabular}{lcccc}
    \toprule
    & \textbf{patt} & \textbf{color} & \textbf{categ} & \textbf{demo} \\
    \midrule
    pattern     & \textbf{0.866} & 0.652 & 0.746 & 0.645 \\
    color       & 0.607 & \textbf{0.912} & 0.601 & 0.603 \\
    category    & 0.720 & 0.623 & \textbf{0.912} & 0.747 \\
    demographic & 0.678 & 0.660 & 0.772 & \textbf{0.899} \\
    \bottomrule
  \end{tabular}
\end{table}

\paragraph{Reference: text-only CLIP baseline.}
The prior text-only slot encoder (CLIP backbone) achieved diagonal mean
AUC $= 0.798$, off-diagonal mean $= 0.704$, and excess leakage $= 0.014$—
also near zero after controlling for dataset co-occurrence.
Table~\ref{tab:auc_clip} shows this matrix for comparison.

\begin{table}[t]
  \centering
  \caption{Linear probe AUC matrix for text-only CLIP slot encoder
           (\emph{prior model, shown for reference}).
           Rows $=$ probe target, cols $=$ feature slot.
           Excess leakage $= 0.704 - 0.690 = 0.014$.}
  \label{tab:auc_clip}
  \small
  \begin{tabular}{lcccc}
    \toprule
    & \textbf{patt} & \textbf{color} & \textbf{categ} & \textbf{demo} \\
    \midrule
    pattern     & \textbf{0.790} & 0.709 & 0.734 & 0.652 \\
    color       & 0.606 & \textbf{0.641} & 0.587 & 0.613 \\
    category    & 0.768 & 0.810 & \textbf{0.904} & 0.753 \\
    demographic & 0.721 & 0.739 & 0.753 & \textbf{0.858} \\
    \bottomrule
  \end{tabular}
\end{table}


\section{Targeted Attribute Intervention}
\label{sec:intervention}

Finally, we test whether the quantized slot codes can be directly edited at query time.

The VQ codebooks in MM-slotgate support direct attribute steering without
re-embedding.
For each query, we choose one slot and replace its quantized code with the
codebook entry most associated with a different target class, using a mapping
computed from training items only.
We then retrieve with the modified representation and measure whether the
top-10 results shift toward the target class while preserving other attributes.

We report \textit{Lift} $=$ HitRate@10 / NullHitRate@10 (prevalence-adjusted
target-class retrieval rate) and \textit{PreserveDelta@10} (mean change in
other-slot match rate; negative values indicate collateral disruption).

\paragraph{MM-slotgate intervention results.}
Table~\ref{tab:intervention} shows results at $n{=}10$.
Color achieves $15.3\times$ lift (Hit $= 0.297$ vs.\ Null $= 0.019$),
a $10.2\times$ improvement over the CLIP-text color lift ($1.5\times$),
consistent with the much stronger color slot encoding (AUC $0.641 \to 0.912$).
Pattern ($9.4\times$), category ($10.7\times$), and demographic ($8.5\times$)
all show strong lifts, with PreserveDelta near zero on non-intervened slots.

\begin{table}[t]
  \centering
  \caption{Codebook intervention results at $n{=}10$ (MM-slotgate).
           PreserveDelta = mean change in other-slot match rate.}
  \label{tab:intervention}
  \small
  \setlength{\tabcolsep}{4pt}
  \begin{tabular}{lrrrr}
    \toprule
    Slot & Hit@10 & Null@10 & Lift & $\Delta$Pres \\
    \midrule
    color       & 0.297 & 0.019 & \textbf{15.3}$\times$ & $-0.026$ \\
    category    & 0.222 & 0.021 & 10.7$\times$ & $-0.042$ \\
    pattern     & 0.099 & 0.011 & 9.4$\times$  & $-0.021$ \\
    demographic & 0.370 & 0.044 & 8.5$\times$  & $-0.080$ \\
    \bottomrule
    \multicolumn{5}{l}{\footnotesize CLIP-text reference: color $1.5\times$,
    category $13.6\times$, pattern $3.0\times$, demo $4.5\times$.}
  \end{tabular}
\end{table}

The largest relative change is on color: lift increases from $1.5\times$ in
the prior text-only slot encoder to $15.3\times$ under MM-slotgate.
This matches the stronger color slot AUC ($0.641 \to 0.912$) and shows that
visual grounding improves both retrieval and codebook-level steering.


\section{Limitations}

The current taxonomy contains only four slots; extending the method
to richer fashion concepts such as occasion, style aesthetic, and brand
remains future work.


\section{Conclusion}

We introduced MM-slotgate, a multimodal slot encoder for controllable
fashion retrieval. Instead of representing each item with a single monolithic
embedding, MM-slotgate factorizes Fashion-CLIP text and image embeddings
into four named attribute slots and lets each slot learn its own text--image
gate. This allows visually grounded attributes such as color and pattern to
draw more from image evidence, while taxonomy-oriented attributes such as
category and demographic can remain more text-driven.

On H\&M, MM-slotgate achieves 0.7566 macro ConstraintSatisfied@10 using
the combined retrieval score, outperforming fCLIP text-only retrieval
(0.4755) and equal-weight MM-global fusion (0.7142). The largest gain is on
color, where CS@10 improves from 0.321 to 0.889. This improvement aligns
with the learned color gate, which assigns 57.4\% weight to image evidence
(\(g_{\text{color}}=0.426\)). More broadly, the learned gates are interpretable
without modality supervision: color leans image, category leans text, and
pattern and demographic lie near the middle.

The same slot structure also preserves controllability. The MM-slotgate slots
show stronger own-slot predictive accuracy than the prior text-only slot
encoder, with diagonal AUC improving from 0.798 to 0.897, while adding no
measured excess leakage beyond the label-correlation baseline. Quantized
slot codes further support direct codebook-level intervention: color achieves
15.3\(\times\) hit-rate lift over unmodified retrieval, compared with 1.5\(\times\)
for the prior text-only slot encoder.

Overall, the results suggest that controllable fashion retrieval benefits from
typed, attribute-conditioned multimodal representations. A single global
text--image blend is not sufficient: different fashion attributes require
different modality mixtures. MM-slotgate provides a simple mechanism for
learning those mixtures while keeping the representation explicitly
addressable for retrieval weighting and targeted intervention.

\bibliographystyle{ACM-Reference-Format}
\bibliography{refs}

\end{document}